\documentclass[journal]{IEEEtran}
\usepackage{caption} 

\usepackage{cite}
\usepackage{lipsum,pifont}
\usepackage{cuted}
\usepackage{bm}
\usepackage{color} 
\usepackage{subfigure}
\usepackage{mathrsfs}
\usepackage{stfloats}
\usepackage{enumitem}

\usepackage{array}

\usepackage{url}
\usepackage{graphicx,amsmath,amssymb,amsfonts,bm}
\usepackage{algorithmic,algorithm}

\usepackage{setspace}

\usepackage{xcolor}
\newtheorem{baseline}{\emph{Baseline}}

\newtheorem{lemma}{\textbf{Lemma}}

\newtheorem{corollary}{\textbf{Corollary}}

\makeatletter
\newcommand{\rmnum}[1]{\romannumeral #1}
\newcommand{\Rmnum}[1]{\expandafter\@slowromancap\romannumeral #1@}
\makeatother
\usepackage{tikz}
\usepackage{framed} 

\usepackage{cancel}

\begin{document}
	% reference control
	\bstctlcite{ref:BSTcontrol}
	
	\title{{\fontsize{22 pt}{\baselineskip}\selectfont Federated Deep Reinforcement Learning for RIS-Assisted Indoor Multi-Robot Communication Systems}}
	
	\author{Ruyu~Luo, Wanli~Ni, \IEEEmembership{Graduate~Student~Member,~IEEE,}~Hui~Tian, \IEEEmembership{Senior~Member,~IEEE,}\\~and~Julian~Cheng, \IEEEmembership{Senior~Member,~IEEE}
		\thanks{Copyright (c) 2015 IEEE. Personal use of this material is permitted. However, permission to use this material for any other purposes must be obtained from the IEEE by sending a request to pubs-permissions@ieee.org.}
		\thanks{This work was supported by the National Natural Science Foundation of China under Grant 62071068. \textit{(Corresponding author: Hui Tian.)}}
		\thanks{R. Luo, W. Ni and H. Tian are with the State Key Laboratory of Networking and Switching Technology, Beijing University of Posts and Telecommunications, Beijing 100876, China (e-mail: luory@bupt.edu.cn; charleswall@bupt.edu.cn; tianhui@bupt.edu.cn).}
		\thanks{J. Cheng is with the School of Engineering, The University of British Columbia, Kelowna, BC V1V 1V7, Canada (email: julian.cheng@ubc.ca).}
	}
	
	\maketitle

	\begin{abstract}
		Indoor multi-robot communications face two key challenges: one is the severe signal strength degradation caused by blockages (e.g., walls) and the other is the dynamic environment caused by robot mobility.
		To address these issues, we consider the reconfigurable intelligent surface (RIS) to overcome the signal blockage and assist the trajectory design among multiple robots.
		Meanwhile, the non-orthogonal multiple access (NOMA) is adopted to cope with the scarcity of spectrum and enhance the connectivity of robots.
		Considering the limited battery capacity of robots, we aim to maximize the energy efficiency by jointly optimizing the transmit power of the access point (AP), the phase shifts of the RIS, and the trajectory of robots.
		A novel federated deep reinforcement learning (F-DRL) approach is developed to solve this challenging problem with one dynamic long-term objective.
		Through each robot planning its path and downlink power, the AP only needs to determine the phase shifts of the RIS, which can significantly save the computation overhead due to the reduced training dimension.
		Simulation results reveal the following findings:
		\rmnum{1}) the proposed F-DRL can reduce at least $86\%$ convergence time compared to the centralized DRL;
		\rmnum{2}) the designed algorithm can adapt to the increasing number of robots;
		\rmnum{3}) compared to traditional OMA-based benchmarks, NOMA-enhanced schemes can achieve higher energy efficiency.
	\end{abstract}
	

	\section{Introduction}
	
	\IEEEPARstart{O}{wing} to their prominent features of flexible deployment and high efficiency, intelligent robots have gained widespread popularity and large-scale implementation in indoor environments, e.g., healthcare surveillance, packet delivery, house cleaning and automatic industrial production \cite{Afrin2021Resource}.
	So far, it is still impractical to deploy all intelligent applications on mobile indoor robots with limited resources such as computing, storage, and batteries \cite{yan2014go}.
	Besides, indoor environment presents several challenges in designing energy-efficient trajectories for robots.
	On the one hand, the line-of-sight (LoS) paths may be severely shields by obstacles that are likely to have non-analytic shapes \cite{Mu2021Intelligent}.
	The resulting signal strength degradation can lead to undesirable effects such as the sudden collision, efficiency reduction and operation restriction. 
	To avoid these potential problems, the reconfigurable intelligent surface (RIS) can be deployed to create a smart propagation environment in an enclosed room\cite{di2020hybrid}, while reducing the hardware cost and system complexity compared with active relays \cite{yang2020intelligent}.
	On the other hand, due to the simultaneous motion of multiple robots, the traditional deterministic strategy is challenging to maintain satisfactory performance of such a highly dynamic system \cite{Park2021Communication}.
	Furthermore, the non-orthogonal multiple access (NOMA) has been deemed as a promising technique for enhancing the robot connectivity and throughput under limited spectrum resources\cite{ni2021federated,Ni2022STAR}.
	By superimposing user signals in different power levels, it is of great significance to jointly optimize the power allocation for interference reduction in NOMA networks\cite{ni2021resource}, while the incorporation of mobile robots and RIS leads to a challenging energy efficiency maximization problem.
	
	%semi-distributed
	Recently, artificial intelligence has played a critical role in realizing smart resource management and automatic network control in 6G networks \cite{yang2020artificial}.
	To deal with the uncertainty and dynamics, deep reinforcement learning (DRL) is acknowledged as a promising method with a high level of intelligence in wireless communications\cite{yang2020deep}.
	%{T}{he} prosperity of machine learning is expected to provide pervasive intelligence for beyond fifth generation (B5G) networks \cite{Chen2019Artificial, Liu2020When}, which will also make the ubiquitous data transmission more sensitive and demanding from privacy protection to resource utilization \cite{Lyu2019Optimal}.
	However, the ever-increasing network scale brings huge communication overhead and unbearable training delay to centralized methods.
	To speed up training and leverage computing capabilities at the network edge, an innovative paradigm is to implement DRL in a federated manner \cite{Nie2021Semi}, which can protect user privacy and alleviate traffic transmission by only exchanging parameters over wireless networks.
	However, the quality of federated training is affected by the channel conditions with all training parameters transmitted over wireless networks, thus the wireless network needs to be reliable over the limited spectrum and power resources\cite{chen2021joint}.
	Meanwhile, the distributed method may obtain a worse solution due to the loss of global information.
	Therefore, it is necessary to develop an intelligent method to maximize energy efficiency in dynamic RIS-assisted wireless systems.
	
	%contributions
	In this paper, we focus on the energy efficiency problem of an RIS-assisted indoor system having multiple mobile robots.
	By jointly optimizing the transmit power at the AP, the phase shifts of the RIS, and the trajectory of robots, a time-coupling resource allocation problem is formulated.
	Considering the trade-off between performance and scalability, a federated deep reinforcement learning (F-DRL) approach is proposed, which can accelerate convergence and is robust to the number of robots.
	To the best of the authors' knowledge, this is the first semi-distributed F-DRL algorithm that combines the centralized RIS configuration with the federated robotic communications.
	The main contributions of this paper can be summarized as follows:
	\begin{enumerate}
		\item
		We incorporate RIS into indoor robot communication systems to overcome signal blockage and avoid motion collision.
		For the maximized energy efficiency of all robots, a non-convex problem is formulated for communication-aware trajectory design.
		The time-coupling and discrete nature make this problem challenging to solve directly.
		\item
		We develop an F-DRL method to optimize the AP transmit power, RIS phase shifts, and robot trajectory in a semi-distributed manner.
		The reduction of control dimension greatly accelerates the convergence at the training stage.
		Benefiting from the decentralized implementation, F-DRL can easily adapt to changes in the robot number.
		\item
		We conduct numerical experiments to show the superiority of the proposed F-DRL.
		Compared to the centralized DRL, our method takes about $86\%$ less training time and is more robust to the dynamic multi-robot environment.
		Simulation results also show that the designed F-DRL can outperform benchmarks in terms of energy efficiency.
	\end{enumerate}

	\section{System Model and Problem Formulation} \label{section_system_model}

	\subsection{System Model} 
\begin{figure}[t]
		\setlength{\abovecaptionskip}{-0.5 mm}
		\centering
		\includegraphics[width=3.5 in]{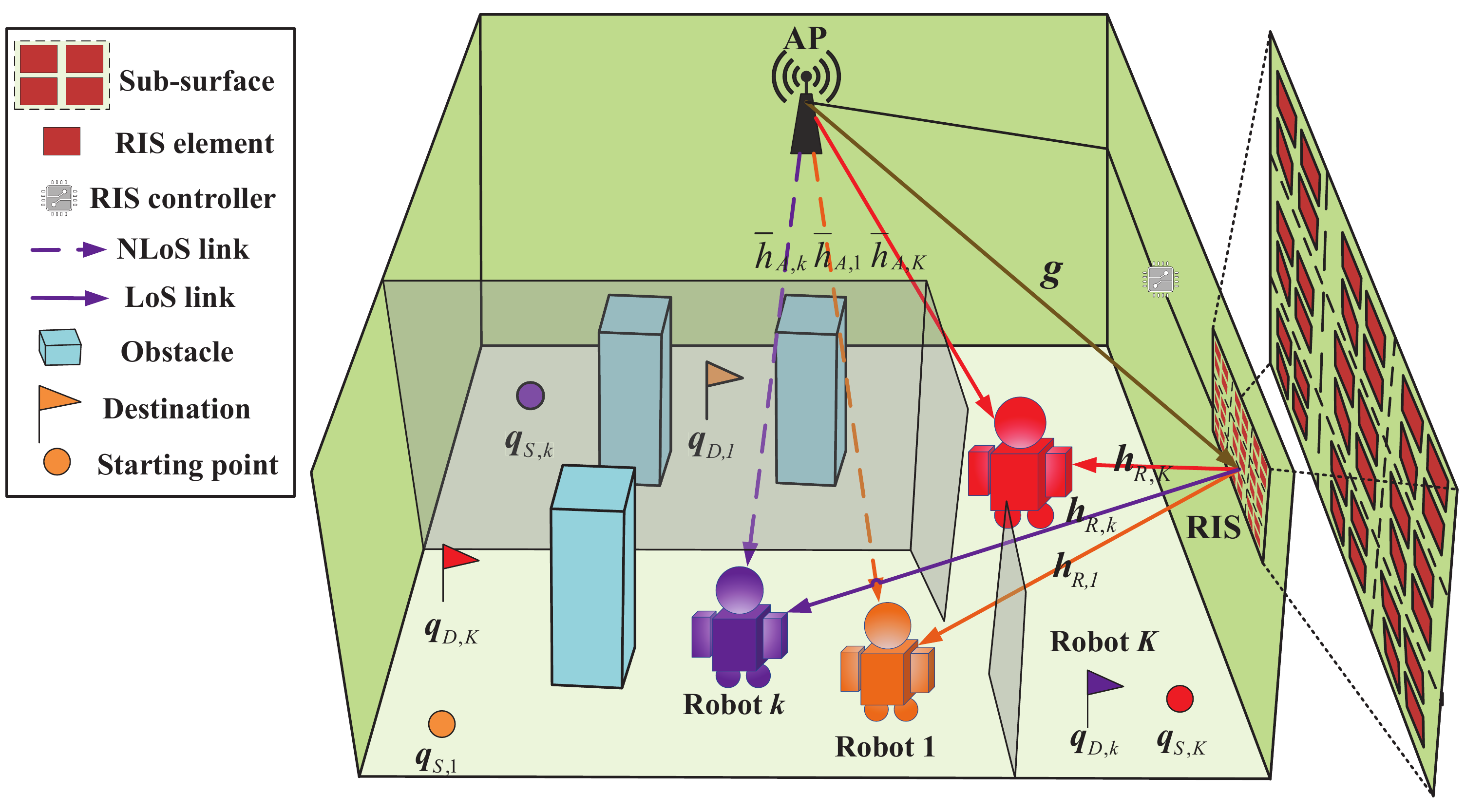}
		\caption{RIS-assisted indoor multi-robot communications}
		\label{fig1}
		\vspace{-3 mm}
	\end{figure}

	As illustrated in Fig. \ref{fig1}, we consider an indoor multi-robot communication system aided by an RIS having $M$ passive reflecting elements.
	Using downlink NOMA techniques, the AP serves $K$ mobile robots\footnote{With the results obtained in this paper, the considered system can be easily extended to multi-antenna cases, which will be included in our future work.}, denoted by $\mathcal{K}=\{1, 2, \dots, K\}$.
	To complete given tasks, we require the $k$-th robot to move from a starting position $\mathbf{q}_{{\rm S},k}$ to a destination $\mathbf{q}_{{\rm D},k}$ within a given deadline $T_{\max}$.
	We define $\mathbf{q}_{k}^t$ as the position of the $k$-th robot at the $t$-th time slot, where $t \in \mathcal{T}_{k}=\{1, 2, \ldots, T_{k}\}$ and $T_{k} \le T_{\max}$ is the total traveling time at the speed $v$.
	For brevity, the time index $t$ is omitted in some parameters.
    We assume that robots update the trajectory each time slot.
	The RIS is divided into ${N}$ sub-surfaces, denoted by $\mathcal{N}=\{1, 2, \ldots, N\}$.
	Let $\theta_{n} \in [0, 2 \pi)$ denote the phase shift of the $n$-th sub-surface.
	Then, the RIS reflection matrix is denoted by $\mathbf{\Theta}={\rm diag}({\bm{\theta}}_{N \times 1} \otimes \mathbf{1}_{(\overline{M/N})\times 1}) = {\rm diag}(e^{j \theta_{1}}, \dots, e^{j \theta_{M}}) \in \mathbb{C}^{M \times M}$ with $M = M_{\rm R} \times M_{\rm R}$, while $M_{\rm R}$ is the element number in the vertical or horizontal direction.
	In view of the hardware implementation, we consider the practical RIS with limited $N_{\rm R} =  2^b$ phase shifts \cite{zhang2020reconfigurable}, where $\theta_{m} \in \mathcal{R} = \{ \frac{1}{2}\Delta_{\rm R},\dots,\frac{2N_{\rm R}-1}{2}\Delta_{\rm R} \}, \forall m$, and $\Delta_{\rm R} = 2 \pi / {N_{\rm R}}$ is the phase resolution\cite{Ni2022Integrating}.

	Let $\bar{h}_{k} \in \mathbb{C}^{1 \times 1} $,  $\mathbf{h}_{k} \in \mathbb{C}^{1 \times M}$ and $\mathbf{g} \in \mathbb{C}^{M \times 1} $ denote the channel coefficients from the AP to the $k$-th robot, from the RIS to the $k$-th robot, from the AP to the RIS, respectively.
	%Assume that the signals reflected by the RIS two or more times are ignored \cite{wu2019intelligent},
	Then the combined channel coefficient experienced by the $k$-th robot is given by $h_k= \mathbf{h}_{k} \mathbf{\Theta} \mathbf{g}+ \bar{h}_{k}$.
	Thus, the received signal at the $k$-th robot is given by
	\begin{equation} \label{eq1}
		\setlength\abovedisplayskip{3pt}
		\setlength\belowdisplayskip{3pt}
		y_k = h_k \sqrt{p_k} s_k + \sum \nolimits_{i \neq k} h_k \sqrt{p_i} s_i + n_k, \ \forall k,
	\end{equation}
	where $s_k$ is the transmit symbol for the $k$-th robot, $p_k > 0$ is the downlink power allocated to the $k$-th robot, and $n_k \sim \mathcal{CN}( 0, \sigma^{2} )$ is the additive white Gaussian noise. 
	
	To alleviate the interference among robots, we apply the successive interference cancellation (SIC) technique.
	Without loss of optimality, the channel coefficients of all robots are ranked by $|h_K| \le \cdots \le |h_2| \le |h_1|$.
	Then, to perform SIC successfully, the transmit power at the AP satisfies the following constraint:
	\begin{equation} \label{power_constraint}
		\setlength\abovedisplayskip{3pt}
		\setlength\belowdisplayskip{3pt}
		\varDelta_{k} = p_{k} \left| h_{k-1}\right| ^{2} - \sum\nolimits_{i=1}^{k-1} p_{i} \left| h_{k-1} \right| ^{2} \ge \rho_{\min}, \ \forall k \ge 2,
	\end{equation}
	where $\rho_{\min} > 0$ is the required gap to distinguish the decoded signal. 
	When the above power constraint is met, the achievable downlink data rate at the $k$-th robot can be obtained by
	\begin{equation} \label{data rate}
		\setlength\abovedisplayskip{3pt}
		\setlength\belowdisplayskip{3pt}
		R_{k} = \log_{2}\left( 1 + \frac{\left| h_k \right| ^{2}p_{k}}{\left| h_k \right| ^{2}\sum_{i=1}^{k-1} p_{i} + \sigma^{2}} \right), \ \forall k.
	\end{equation}
	
	Since the energy consumed by motion is much larger than that consumed by communication, this paper mainly focuses on the motion energy cost.
	Therefore, the total motion energy consumed by the $k$-th robot is expressed as \cite{yan2014go}
	\begin{equation}
		\setlength\abovedisplayskip{3pt}
		\setlength\belowdisplayskip{3pt}
		E_{k} = E_1 T_k v + E_2 T_k, \ \forall k,
	\end{equation} 
	where $E_1$ and $E_2$ are two constants related to the mechanical output power and the transforming loss, respectively \cite{mei2006deployment}.
	Their values depend on the exact robot motion model.

	\subsection{Problem Formulation}
	By optimizing the transmit power at the AP, the phase shifts of the RIS, and the trajectory of robots, this paper aims to maximize the total energy efficiency of all robots during the mission.
	Subject to the constraints of transmit power, phase shifts and robot mobility, a long-term optimization problem is formulated as
	\begin{subequations} \label{sumrate_maximization}
		\setlength\abovedisplayskip{3pt}
		\setlength\belowdisplayskip{3pt}
		\begin{eqnarray}          \label{sumrate_maximization_objective}
			&\max \limits_{\mathbf{\Theta}, \bm{Q}, \bm{p}} & \frac{1}{T_{k}}\sum\nolimits_{t=1}^{T_k} \sum\nolimits_{k=1}^{K} \frac{R_{k}^{t}}{E_{k}} \\
			\label{constraint1}
			& {\rm s.t.}&
			\mathbf{q}_{k}^1 = \mathbf{q}_{{\rm S}, k}, \ \mathbf{q}_{k}^{T_{k}} = \mathbf{q}_{{\rm D}, k}, \ \forall k, \\ 
			\label{constraint6}
			& {} & 
			|h_K^t| \le \cdots \le |h_2^t| \le |h_1^t|, \ \forall t, \\ 
			\label{constraint2}
			& {} & 	\varDelta_k^t \ge \rho_{\min}, \ p_{k}^{t} > 0, \ \forall k, \forall t, \\
			\label{constraint3}
			& {} & x_{\min} \le x_{k}^{t} \le x_{\max}, \ \forall k, \forall t,\\ 
			\label{constraint4}
			& {} & y_{\min} \le y_{k}^{t} \le y_{\max}, \ \forall k, \forall t,\\ 
			\label{constraint7}
			& {} & \sum\nolimits_{k=1}^{K} p_{k}^{t} \le P_{\max}, \ \forall t, \\
			\label{constraint5}
			& {} & 
			\theta_{n}^{t} \in \mathcal{R}, \ \forall n, {\forall t,}		
		\end{eqnarray}
	\end{subequations}	
	where $\bm{Q} = [\mathbf{q}_{1}, \mathbf{q}_{2}, \dots, \mathbf{q}_{K}]^{\rm T}$ denotes the trajectory design of all robots and $\bm{p} = [p_{1}, p_2, \dots, p_{K}]^{\rm T}$ is the power allocation strategy at the AP.
	However, the formulated problem (\ref{sumrate_maximization}) is difficult to be solved by existing optimization methods and is also challenging to be solved optimally, due to the following reasons.
	First, multiple optimization variables, $\{ \mathbf{\Theta}, \bm{Q}, \bm{p} \}$, are closely coupled in the objective function (\ref{sumrate_maximization_objective}).
	Second, the achievable data rate $R_{k}^{t}$ is not a continuous function due to the discrete phase shifts and the position-dependent channel coefficients.
	Third, the simultaneous motion of multiple robots also makes problem (\ref{sumrate_maximization}) hard to solve even if only the subproblem of trajectory design is considered.
	To sum up, traditional one-shot optimization methods do not apply to this dynamic problem with a time-coupling objective.
	Thus, it is necessary to develop an intelligent method to address this challenging problem in an efficient manner.

	\section{Proposed F-DRL Approach}\label{DRL BASED ALGORITHM}
	
\begin{figure}[t]
		\setlength{\abovecaptionskip}{-0.5 mm}
		\centering
		\includegraphics[width=3.5 in]{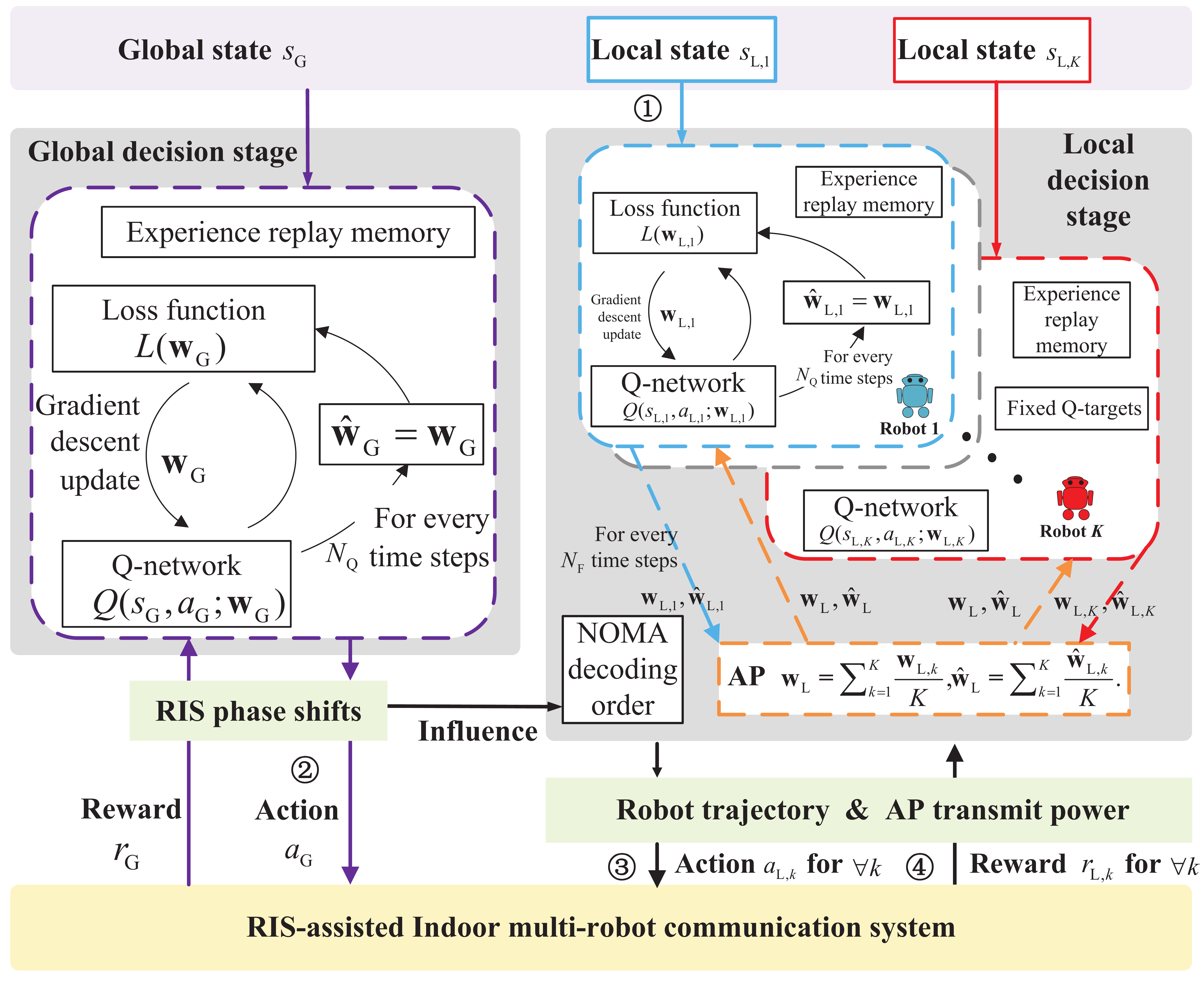}
		\caption{Proposed F-DRL approach for communication-aware trajectory design}
		%	\caption{Proposed F-DRL approach with five steps, including \textcircled{1} input states, \textcircled{2} output RIS phase shifts, \textcircled{3} decide NOMA decoding order, \textcircled{4} output robot trajectory and AP transmit power, \textcircled{5} return rewards.}
		\label{fig2}
		\vspace{-3 mm}
	\end{figure}

	In this section, we develop an F-DRL approach that is capable of accelerating the training process and obtaining high performance in terms of energy efficiency.
	%	Compared to existing DRL methods, the proposed F-DRL can adapt to the changes in the number of robots. %{\color{blue}{while obtaining the a high-performance solution in RIS design.}}
	As shown in Fig. \ref{fig2}, the F-DRL approach is split into two stages: the global stage for RIS configuration and the local stage for joint robot trajectory and transmit power control. 	
	
	\vspace{-4 mm}
	\subsection{Global Decision Stage}\label{algorithm design}
	At the global decision stage, the AP adjusts the RIS configuration with global state information.
	Specifically, we define the phase shift design problem as a Markov decision process (MDP), denoted by a transition tuple having three elements: $\left\langle \mathcal{S}_{\rm G}, \mathcal{A}_{\rm G}, \mathcal{R}_{\rm G} \right\rangle$, where $\mathcal{S}_{\rm G}$ is the state space, $\mathcal{A}_{\rm G}$ is the action space, and $\mathcal{R}_{\rm G}$ is the reward.
	\begin{itemize}
		\item
		\textbf{\emph{State space:}}
		Let $s_{\rm G}^{t} \in \mathcal{S}_{\rm G}, \forall t$. Since the combined channel coefficients $\left(h_k\right) _{k \in \mathcal{K}}$ remain unknown before the RIS phase shifts are designed, the coefficients of AP-robot links $\left( \bar{h}_{k} \right)_{k \in \mathcal{K}}$ are considered as the channel features. Thus, the global state is defined as
		\begin{equation} \label{state_space}
			\setlength\abovedisplayskip{3pt}
			\setlength\belowdisplayskip{3pt}
			s_{\rm G}^{t}=\left\{ \left( \mathbf{q}_{k}^{t}, \bar{h}^{t}_{k} \right) | \ { \forall k \in \mathcal{K}} \right\}, \ \forall t,
		\end{equation}
		where the position $\mathbf{q}_{k}^{t}$ can be obtained by the simultaneous localization and mapping algorithm\cite{gao2021slarm}.
		Meanwhile, the continuous 2D map is discretized into grids with the length of $\Delta_{\rm S}$, while sampling positions are in the center of each grid and satisfy constraints in (\ref{constraint3}) and (\ref{constraint4}).\footnote{Using the default track curve \cite{rau2020generating}, the discrete sampling points can be reconstructed into continuous curves.}
		\item 
		\textbf{\emph{Action space:}}
		Let $a_{\rm G}^{t} \in \mathcal{A}_{\rm G}, \forall t$. Then, the action for RIS phase shift design is defined as
		\begin{equation} \label{action_space}
			\setlength\abovedisplayskip{3pt}
			\setlength\belowdisplayskip{3pt}
			a_{\rm G}^{t} = \left\{ \theta_{n}^{t} | \ \forall n \in \mathcal{N} \right\}, \ \forall t, 
		\end{equation}
		where $\theta_{n} \in \mathcal{R}$ is the discrete phase shift adopted by the $n$-th RIS sub-surface.
		\item
		\textbf{\emph{Reward:}}
		With the aim of maximizing the sum rate, the reward is defined as
		\begin{equation} \label{reward_function}
			\setlength\abovedisplayskip{1pt}
			\setlength\belowdisplayskip{3pt}
			r_{\rm G}^{t} = \tau_{1} \sum\nolimits_{k=1}^{K} R_{k}^{t}, \ \forall t,
		\end{equation}
		where $\tau_1$ is a constant.
		Let $r_{\rm G}^{t} < 0$ to avoid robots wondering.
		Additionally, it is inappropriate to put the sum of combined channel coefficients $\left( h_k \right)_{ k \in \mathcal{K} }$ into the reward function, because it is necessary for NOMA to maintain the distinctness among different signals.
	\end{itemize}
	
	\vspace{-4 mm}
	\subsection{Local Decision Stage}
	\vspace{-0.5 mm}
	At the local decision stage, each robot determines its trajectory and downlink transmit power with local state information. 
	Because the control dimension of centralized DRL multiplies with the increase of robots, we propose to train robots locally and then aggregate a global model in a federated manner.
	The MDP of local transition tuple $\left\langle \mathcal{S}_{{\rm L},k}, \mathcal{A}_{{\rm L},k}, \mathcal{R}_{{\rm L},k} \right\rangle$ maintained by the $k$-th robot is defined as follows.
	\begin{itemize}
		\item
		\textbf{\emph{State space:}}
		Let $s_{{\rm L},k}^{t} \in \mathcal{S}_{{\rm L},k}, \forall t$. Then, the local state is defined as
		\begin{equation} \label{state_space2}
			\setlength\abovedisplayskip{1pt}
			\setlength\belowdisplayskip{3pt}
			s_{{\rm L},k}^{t} = \left\{\mathbf{q}_{k}^{t}, \bar{{h}}_{A,k}^{\ t} \right\}, \ \forall k, \forall t,
		\end{equation}
		where the local state $s_{{\rm L},k}^{t}$ is a part of the global state $s_{\rm G}^{t}$.
		\item
		\textbf{\emph{Action space:}}
		Let $a_{{\rm L},k}^{t}\in \mathcal{A}_{{\rm L},k}, \forall k, \forall t$. Then, the local action for trajectory design and transmit power control is defined as
		\begin{equation} \label{action_space2}
			\setlength\abovedisplayskip{1pt}
			\setlength\belowdisplayskip{3pt}
			a_{{\rm L},k}^{t} = \left\{ o_{k}^{t}, p_{k}^{t} \right\}, \ \forall k, \forall t,
		\end{equation}
		where the $k$-th robot orientation $o_{k} \in \{ n, s, e, w \}$ intends that robots move in four directions, i.e., north, south, east or west.
		To satisfy constraints in (\ref{constraint6}), (\ref{constraint2}) and (\ref{constraint7}), the first robot must guarantee $p_1 < {P_{\max}}/{2^{K-1}}$.
		Inspired by the discrete power control, we have $p_{k} \in \{ P_{\max}/2, \dots, {P_{\max}/{ 2^{N_{\rm P}}} } \}$ and $N_{\rm P} \ge K$.
		\item 
		\textbf{\emph{Reward:}}
		{To maximize the energy efficiency, we define the local reward $r_{{\rm L},k}^{t}$ as}
		\begin{equation} \label{reward_function2}
			\setlength\abovedisplayskip{3pt}
			\setlength\belowdisplayskip{3pt}
			r_{{\rm L},k}^{t} = \phi R_{k}^{t} + \psi  R_{{\rm D},k}^{t} + R_{\rm time} + R_{\rm goal}, \ \forall k, \forall t, 
		\end{equation}
		where the guidance reward $R_{{\rm D},k}^{t} = d_{{\rm D},k}^{ \ t - 1} - d_{{\rm D},k}^{ \ t}$ for $t \ge 2$ and $d_{{\rm D},k}^{\ t}$ is the distance between the $k$-th robot and its destination at the $t$-th time slot.
		The guidance reward $R_{{\rm D},k}^{t}$ leads the $k$-th robot to reach its destination.
		Moreover, the time cost $R_{\rm time}$ is a constant and $R_{\rm time} < 0$.
		If the $k$-th robot arrives at its destination, it will gain a positive reward $R_{\rm goal}${; otherwise we have} $R_{\rm goal} = 0$.
		In this paper, the parameter $\phi$ must guarantee $R_{\rm time} + \phi R_{k}^t < 0$ in most cases to prevent robots from wandering.
	\end{itemize}

	\begin{algorithm}[t]
		\caption{Proposed F-DRL Approach}
		\label{F-DRL}
		\begin{algorithmic}[1]
			\renewcommand{\algorithmicrequire}{\textbf{Initialize}}
			\renewcommand{\algorithmicensure}{\textbf{Output}}
			\STATE \textbf{Initialize} the environment $E$ and DQN agents.
			\FOR{episode $e = 1 : N_{e}$}
			\FOR{time step $t = 1 : T_{e}$}
			\STATE Select RIS phase shifts $a_{\rm G}^{t}$ with $Q_{\rm G}( s_{\rm G}^{t}, a_{\rm G}^{t}; \mathbf{w}_{\rm G}^{t})$.
			\STATE Robot $k \!\!\in\!\! \mathcal{K}\!$ selects $a_{{\rm L},k}$ with $\!Q_{{\rm L},k}(s^{t}_{{\rm L},k}, a^{t}_{{\rm L},k}; \mathbf{w}^{t}_{{\rm L},k}\!)\!$.
			\FOR{DQN agent $i \in \mathcal{K} \cup \{ k_{\rm G} \}$}
			\STATE Execute action, get reward and reach next state.
			\STATE Store the transition in reply memory $\mathcal{D}_i$.
			\STATE Sample random mini-batch $\mathcal{D}_{i,{\rm 0}}$ from $\mathcal{D}_i$.
			\STATE Perform the gradient descent step to update $\mathbf{w}_i^t$.
			\STATE Reset $\hat{\mathbf{w}}_i^t = \mathbf{w}_i^t$ every $\mathcal{N}_{\rm Q}$ time steps.
			\ENDFOR
			\FOR{each robot $k \in \mathcal{K}$}
			\STATE Upload $\mathbf{w}_{{\rm L},k}^t, \!\!\hat{\mathbf{w}}_{{\rm L},k}^{t}$ to the AP every $\!\mathcal{N}_{\rm F}\!$ time steps.
			\IF{receive new global weights  $\mathbf{w}_{\rm L}^{t},\hat{\mathbf{w}}_{\rm L}^{t}$} 
			\STATE Download weights $\mathbf{w}_{{\rm L},k}^t=\mathbf{w}_{\rm L}^t$,$\hat{\mathbf{w}}_{{\rm L},k}^t=\hat{\mathbf{w}}_{\rm L}^t$.
			\ENDIF
			\ENDFOR
			\STATE AP aggregates global weights every $\mathcal{N}_{\rm F}$ time steps.
			\ENDFOR
			\ENDFOR
		\end{algorithmic}
	\end{algorithm}

	\subsection{Global Aggregation}
	Take training deep Q-network (DQN) as an example. All agents collaboratively build a shared DQN, where the replay memory $\mathcal{D}$ and $\epsilon$-greedy policy are considered.
	For each DQN agent $i \in \mathcal{K} \cup \{ k_{\rm G} \}$, the online Q-network and the target Q-network are defined as  $Q (s_{i}^{t}, a_{i}^{t}; \mathbf{w}_{i}^{t})$ and $Q(s_{i}^{t}, a_{i}^{t}; \hat{\mathbf{w}}_{i}^{t})$, respectively.
	To update the online Q-network, each agent performs the gradient descent step with a learning rate $\alpha>0$ on the loss function.
	Meanwhile, the target Q-network reset $\hat{\mathbf{w}}_{i}^t = \mathbf{w}_{i}^t$ every $\mathcal{N}_{\rm Q}$ time steps.

	Besides, the $k$-th robot trains networks locally and uploads relevant weights $\mathbf{w}_{{\rm L},k}$, $\hat{\mathbf{w}}_{{\rm L},k}$ every $\mathcal{N}_{F}$ time steps during the local decision stage. 
	At each aggregation step, all robots upload local weights to the AP at the $t$-th time slot, and the AP aggregates the global weights $\mathbf{w}_{\rm L}^{t}$ and $\hat{ \mathbf{w} }_{\rm L}^t$ as
	\begin{equation} \label{FL_UPDATE}
		\setlength\abovedisplayskip{3pt}
		\setlength\belowdisplayskip{3pt}
		\mathbf{{w}}_{\rm L}^{t} = \frac{1}{\mathnormal{K}} \sum\nolimits_{k=1}^{\mathnormal{K}} {\mathbf{{w}}_{{\rm L},k}^t},\ 
		\hat{\mathbf{{w}}}_{\rm L}^{t} = \frac{1}{\mathnormal{K}} \sum\nolimits_{k=1}^{\mathnormal{K}} {\hat{\mathbf{{w}}}_{{\rm L},k}^{t}}, \ \forall k, \forall t.
	\end{equation}
	Then, the updated global weights are sent back to local robots at the next time step until convergence.
	
	%Moreover, benefiting from the semi-distributed structure and federated aggregation, the proposed F-DRL can easily adapt to the ever-increasing robots and design the system with suitable knowledge in above stages.
	Compared to traditional optimization algorithms, the proposed intelligent approach can adapt to the uncertainty and dynamics of indoor systems.
	Moreover, due to the semi-distributed training and decentralized execution, the proposed F-DRL approach can significantly reduce the communication overhead and effectively alleviate privacy leakage.
	
	\subsubsection{Overall Training Methodology}
	
	As shown in Fig. \ref{fig2}, the proposed F-DRL approach has four steps.
	(1) State observation: agents observe the environmental states. 
	(2) RIS action execution: the AP controls RIS phase shifts $a_{\rm G}^{t}$ according to $Q_{\rm G}(s_{\rm G}, a_{\rm G}; \mathbf{w}_{\rm G})$ obtained at the global decision stage, and determines the NOMA decoding order.
	(3) Robot action execution: the $k$-th robot decides its action $a_{{\rm L},k}^{t}$ of the orientation and downlink transmit power based on $Q_{{\rm L},k}(s_{{\rm L},k}, a_{{\rm L},k}; \mathbf{w}_{{\rm L},k})$.
	(4) Experience storage: agents obtain rewards and store transitions. 
	Algorithm \ref{F-DRL} shows the detailed training procedure of the proposed F-DRL approach.
	On account of the interaction between the local agents and the global agent, the proposed F-DRL approach operates in a semi-distributed manner.

	\subsubsection{Complexity Analysis} \label{Complexity Analysis}
	By reducing the control dimension, the complexity of F-DRL is lower than that of centralized learning.
	More precisely, the complexity for DQN using one-dimensional replay memory is $\mathcal{O}(1)$.
	The computational complexity of each agent mainly depends on the transition and back-propagation, which can be calculated by $\mathcal{O} \left( |\mathcal{D}| + abE |\mathcal{D}_{0}|\right) $, where $a$, $b$ and $E$ denote the number of layers, the transitions in each layer and the number of episodes, respectively.
	Moreover, the action space size of F-DRL at the global and local decision stage are $(N_{\rm R})^{N}$ and $(4N_{\rm P})^{K}$, respectively, but that of centralized DRL is $(4N_{\rm P})^{K} \times (N_{\rm R})^{N}$.
	Therefore, the proposed F-DRL has a lower complexity as compared to centralized DRL.
	The theoretical analysis of F-DRL convergence has been completed in \cite{wang2020federated}.
	A detailed proof is omitted here for brevity.
	In the following, we conduct experiments to show the convergence behavior of F-DRL.
	%{\footnote{\color{blue}{The convergence of the proposed semi-distributed algorithm is challenging, while the convergence of centralized/distributed DQN can be guaranteed \cite{wang2020federated}.}}}
	%The convergence of the proposed semi-distributed algorithm is challenging due to the coupled two stages that conclude different DQN structures assisted by federated learning, while the convergence of dependent DQN and distributed DRL can be guaranteed.
	
	\begin{table}[t]
		\caption{Parameter Settings}
		\label{table1}  
		\centering  
		\begin{tabular}{|c|c|c|c|}
			\hline
			Parameter & Value & Parameter & Value\\
			\hline
			$\Delta_{\rm S}$ & $0.5 \ {\rm m}$ & $\alpha$ & $0.0001$ \\
			\hline
			$|\mathcal{D}_{\rm 0}|$ & $128$ &  $\mathcal{N}_{\rm F}$ & $25$\\
			\hline
			$R_{\rm time}$ & $-1$ & $R_{\rm goal}$ & $\{ 0, 100 \}$\\
			\hline
			$\tau_{1}$ & $0.1$ &  $v$ & $0.5 \ {\rm m/s}$\\
			\hline
			$E_1$& $7.4$ &  $E_2$& $0.29$\\
			\hline
		\end{tabular}
		\vspace{-0 mm}
	\end{table}
	
	\begin{figure}[t]
		\centering
		\includegraphics[width=3.5in]{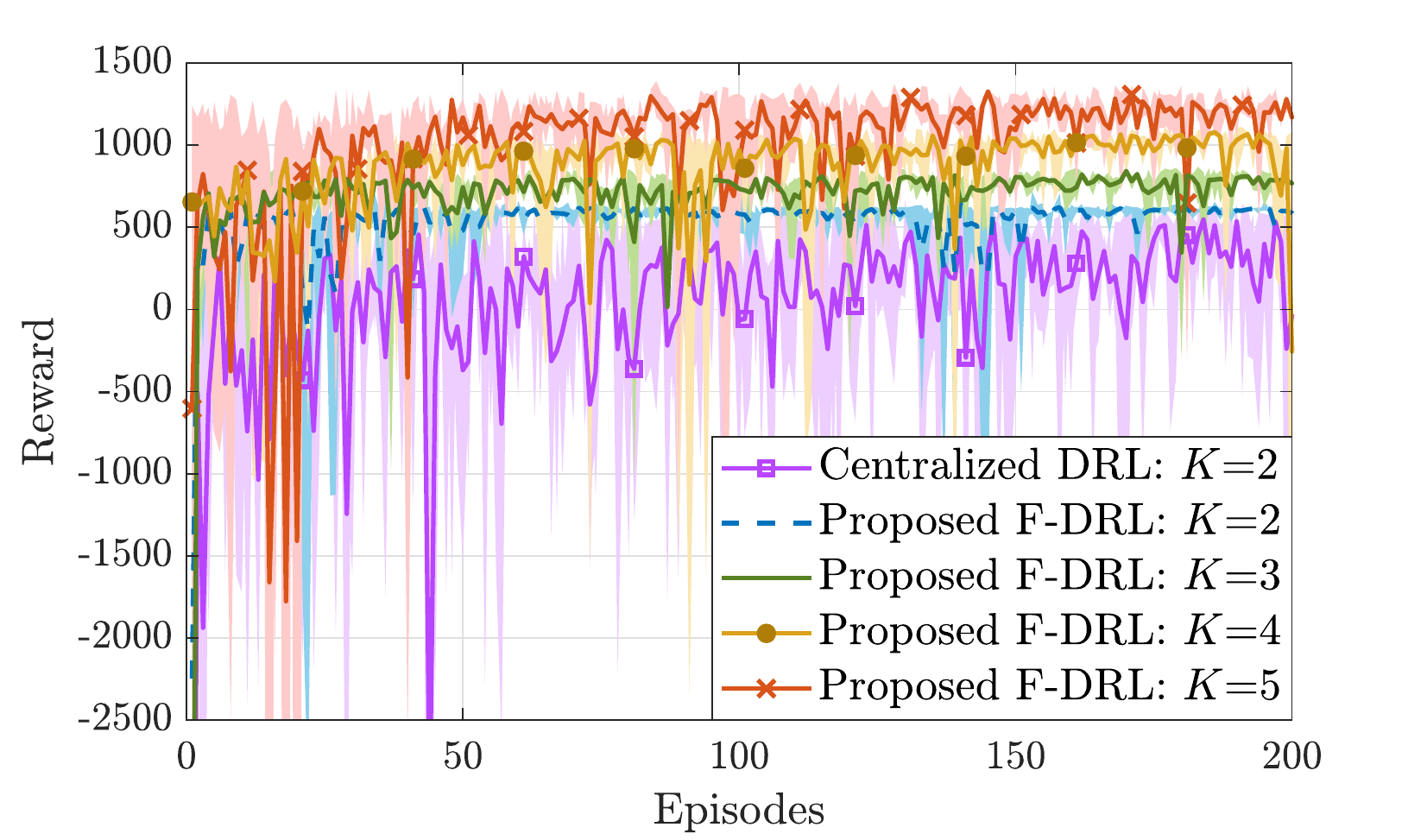}
		\caption{Convergence comparison versus episodes}
		\label{fig3}
		\vspace{-0 mm}
	\end{figure}
	
	\begin{figure*}[t]
		\centering
		\hspace{0.1cm} 
		\subfigure[$M_{\rm R}=0$]{
			\includegraphics[width=3.5in]{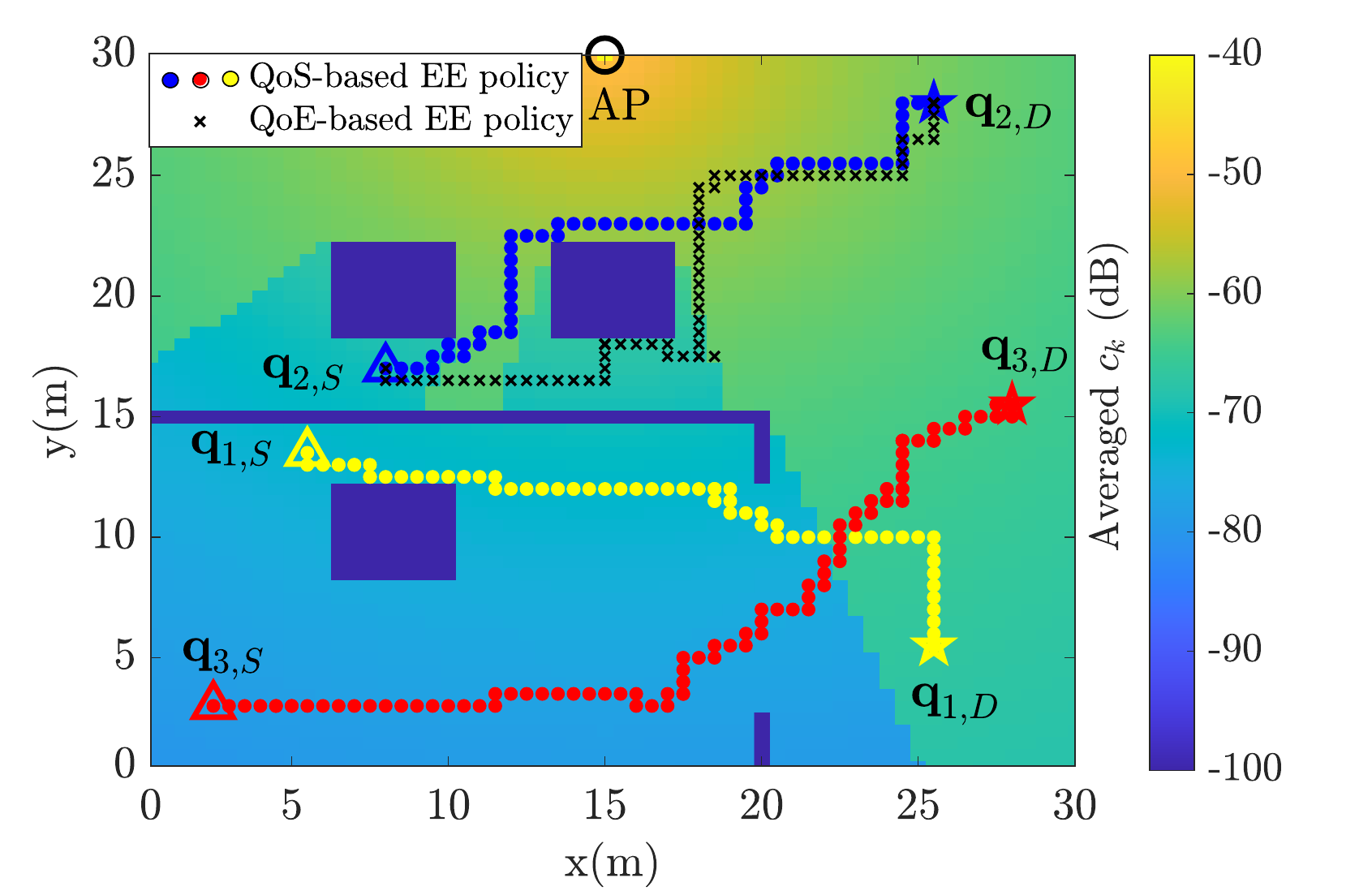}
		}\hspace{-0.5cm}
		\subfigure[$M_{\rm R}=20$]{
			\includegraphics[width=3.5in]{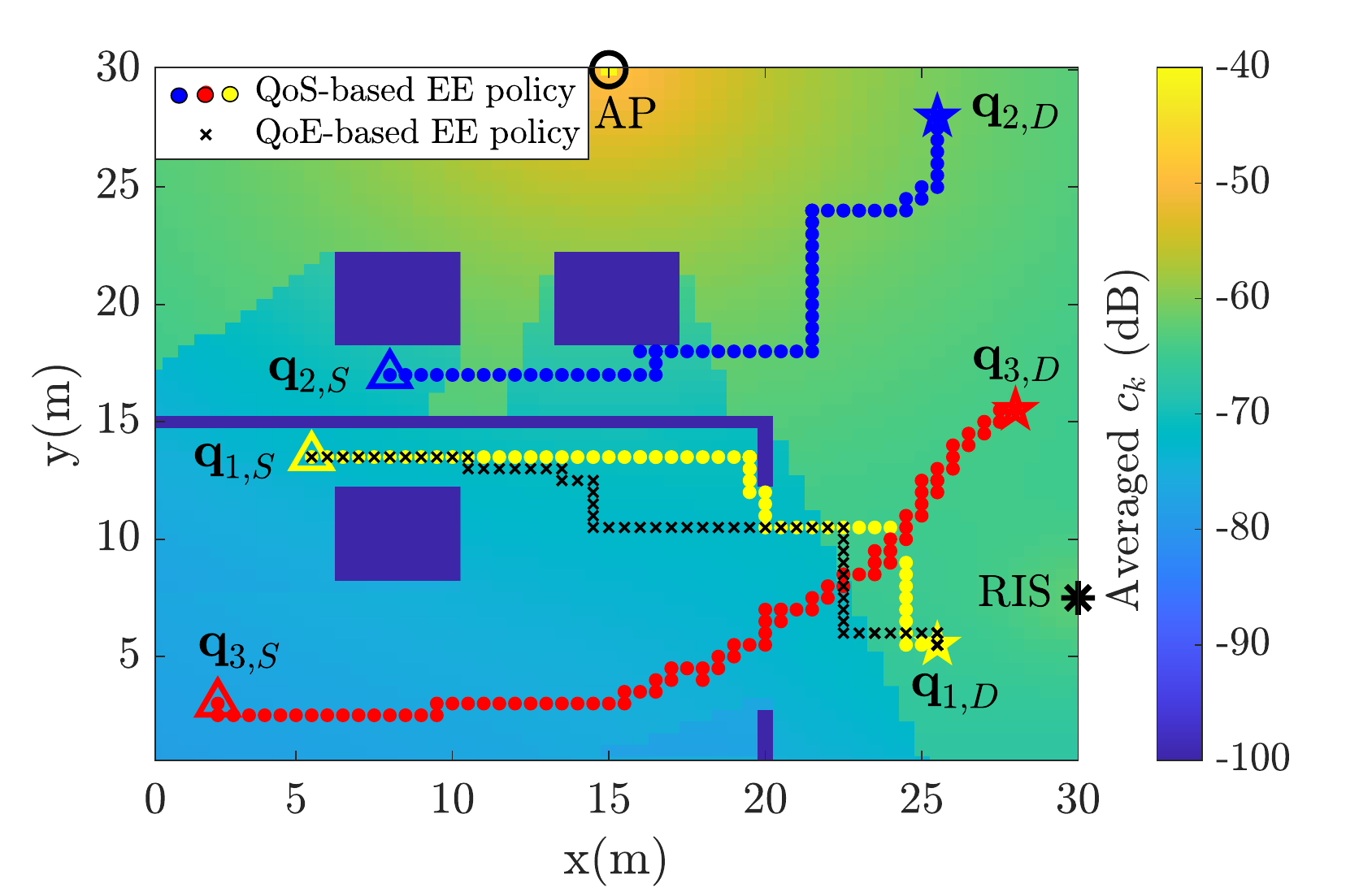}
		}\hspace{-0.5cm} \\
		\subfigure[$M_{\rm R}=30$]{
			\includegraphics[width=3.5in]{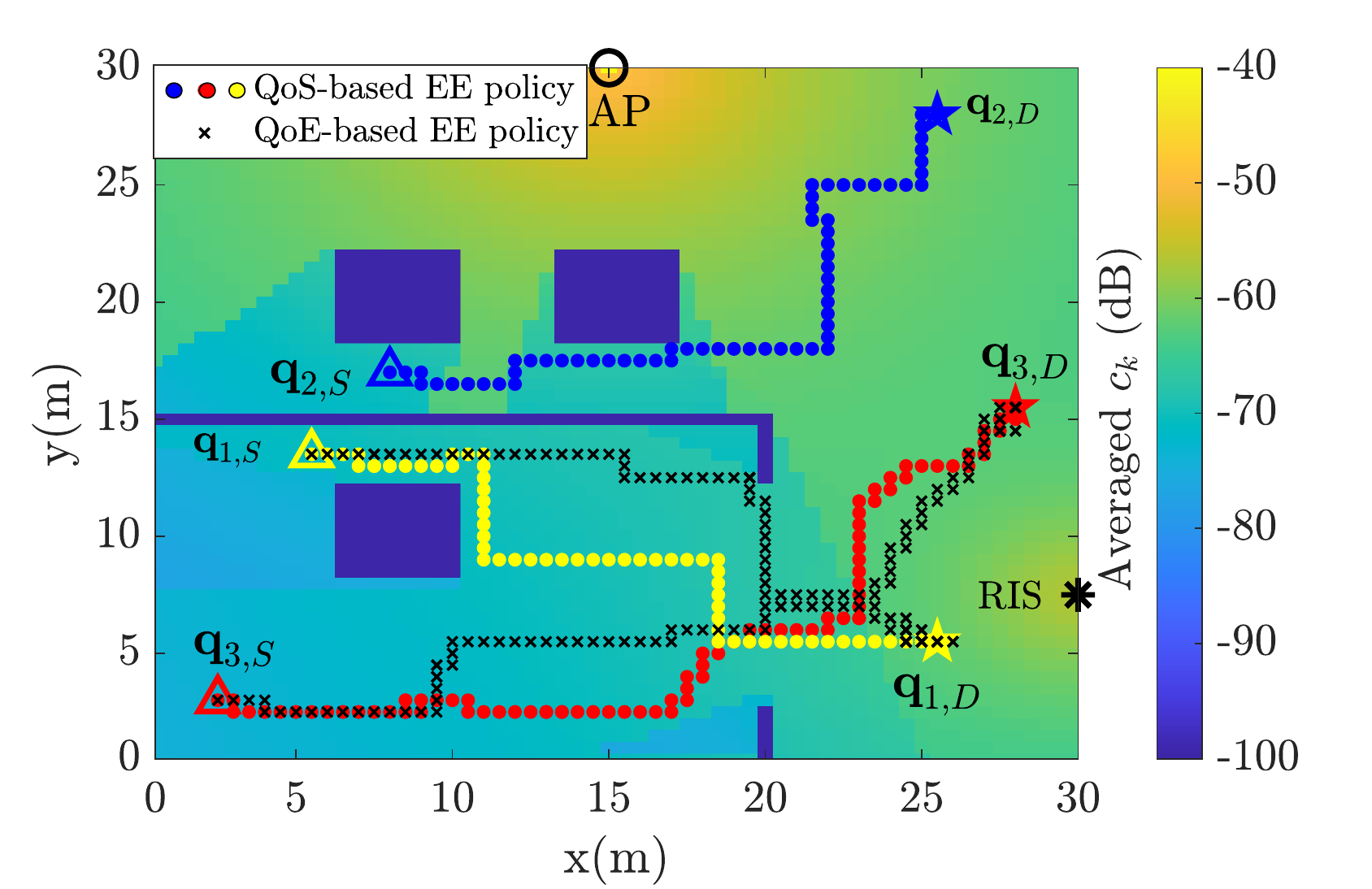}
		}\hspace{-0.5cm}
		\subfigure[$M_{\rm R}=40$]{
			\includegraphics[width=3.5in]{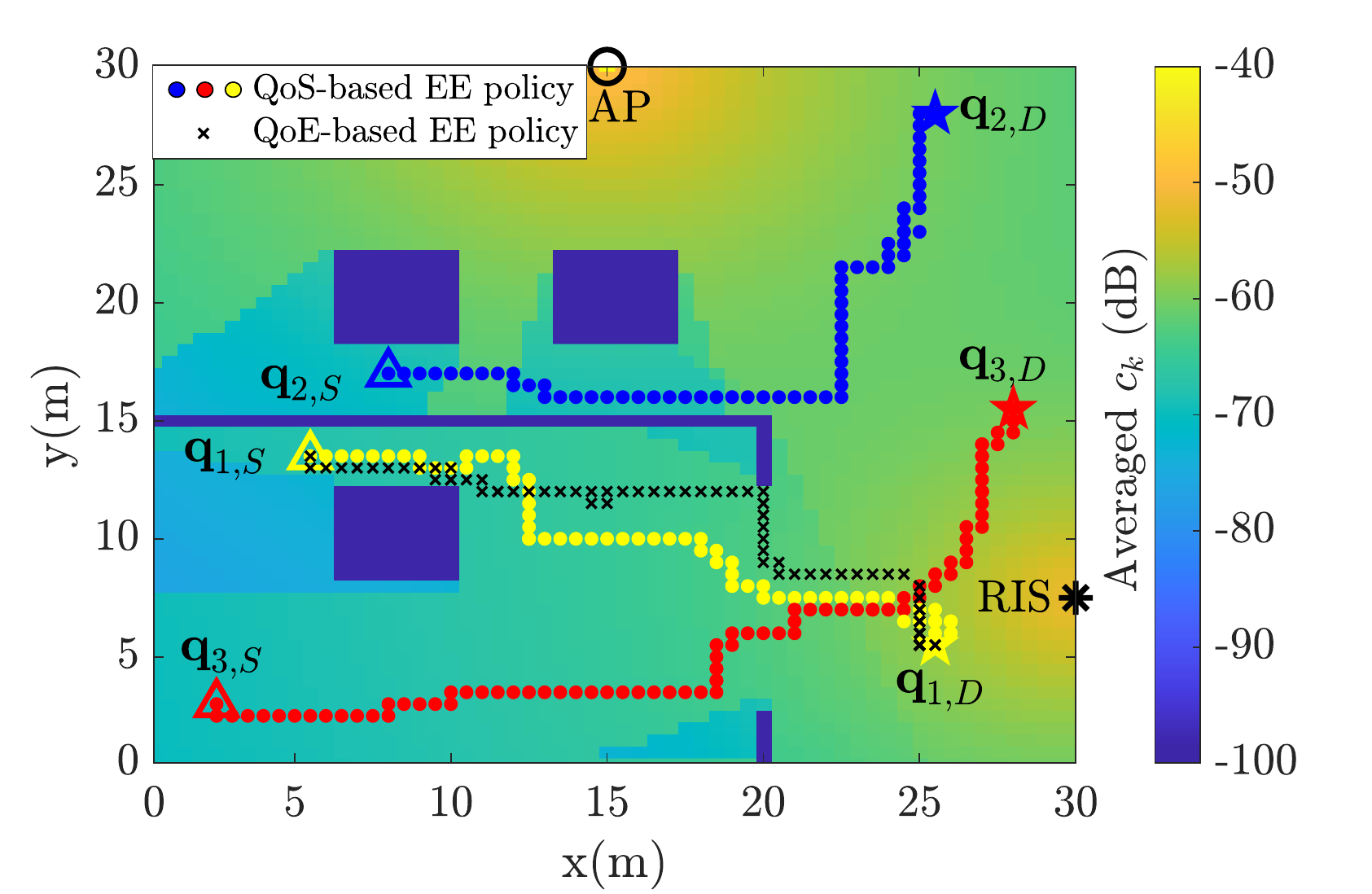}
		}\hspace{-0.5cm}%\vspace{-0.2cm}
		\caption{Trajectory of robots under different values of $M_{\rm R}$, where the red, blue and yellow points denote the robot trajectory using QoS-based energy efficiency (EE) policy, and the black markers denote the trajectory using QoE-based EE policy.}
		\label{fig5}
		
	\end{figure*}
	
	\begin{figure}[t]
		\vspace{-0 mm}
		\centering
		\includegraphics[width=3.5in]{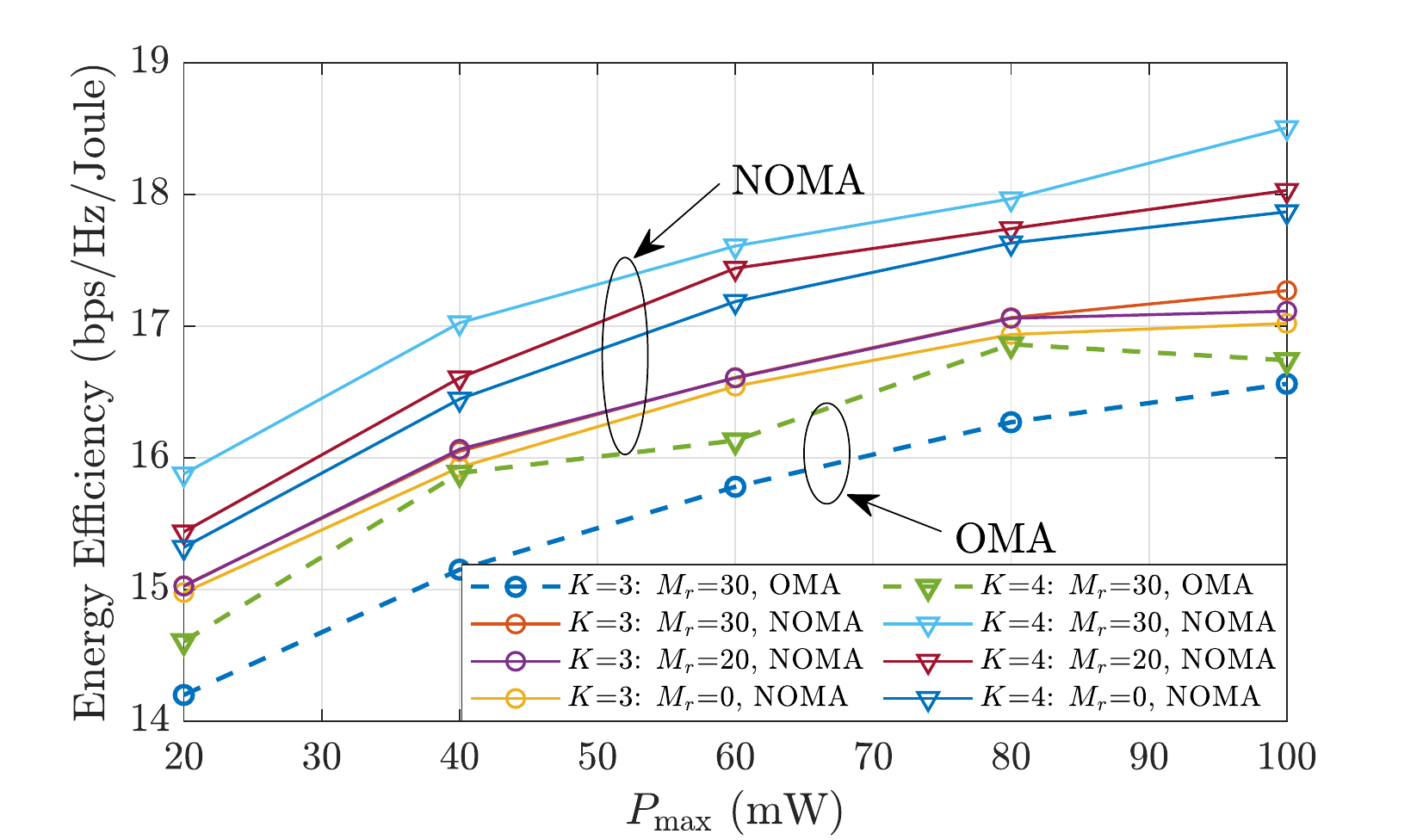}
		\caption{Energy efficiency versus power budget $P_{\max}$}
		\label{fig4}
		\vspace{-3 mm}
	\end{figure}

	%\newpage
	
	\section{Numerical Results}
	
	In this section, we verify the efficiency and robustness of the proposed F-DRL approach for the considered communication system.
	In the simulation, the robots are randomly located, while the AP and the RIS are located at $(15, 30, 2)$ and $(30, 7.5, 2)$, respectively.
	The maximum transmit power of the AP is $P_{\max}= 20 \ {\rm dBm}$ and the noise power spectral density is $N_0=-100 \ {\rm dBm/Hz}$.
	The channel model is the same as the settings in \cite{luo2021communication}.
	Other parameters are given in Table \ref{table1}.
	For comparison, we consider the following baselines:
	
	\begin{itemize}
		\item 
		\begin{baseline}[Centralized DRL \cite{Nguyen2020Deep}]\label{baseline1}
			All decisions of problem (\ref{sumrate_maximization}) are output by a centralized DQN.
			The global state is defined as $s_{\rm DQN}^{t} = s_{\rm G}^{t}$,
			the global action is $a_{\rm DQN}^{t} = \{ \mathbf{\Theta}^{t}, ( o_{k}^{t}, p_{k}^{t} ) | \ \forall k \in \mathcal{K} \}$,
			and the reward is set as $r_{\rm DQN}^{t} = \sum_{k=1}^{K}r_{{\rm L},k}^{t} + {r_{\rm G}^{t}}/{10}$ to prevent robot wandering.
		\end{baseline}
		\item
		\begin{baseline}[OMA-RIS-based scheme \cite{Mu2021Intelligent}]\label{baseline2}
			In this scheme, the orthogonal multiple access (OMA) is considered between the robots and the AP. The entire bandwidth is equally divided by robots, and $R_{k} = {1}/{K}\log_{2}(1+\frac{| h_k| ^{2}p_{k}}{{\sigma^{2}}/{K}})$ is the downlink data rate of the $k$-th robot.
		\end{baseline}
		\item
		\begin{baseline}[QoE-based energy efficiency policy \cite{Liu2021RIS}]\label{baseline3}
			Using the quality of experience (QoE) metric to evaluate the performance of each robot, we have $\eta_{k} = C_{1}\lg(R_{k} ) + C_{2}$, where $C_{1}$ and $C_{2}$ are constants. Meanwhile, we replace $R_{k}$ with $\eta_{k}$ in the reward returned back to each agent. 
		\end{baseline}
	\end{itemize}

	In Fig. \ref{fig3}, the convergence performance of the proposed F-DRL is shown, where the total rewards versus training episodes under different schemes are compared.
	We consider the system with $M_{\rm R} = 30, N_{\rm R} = 4, N_{\rm P} = 6, N = 1$ but different $K$.
	When $K = 2$, we find the proposed F-DRL takes at least $86\%$ less training time than Baseline \ref{baseline1}. 
	More significantly, the performance gain of the proposed F-DRL grows with the increase of $K$, while Baseline \ref{baseline1} cannot work when $K > 3$.
	This is due to the fact that the global action $s_{\rm DQN}^{t}$ increases exponentially with $K$.
	In contrast, F-DRL is robust to the changes in the number of robots.
	On the whole, compared with Baseline \ref{baseline1}, one can observe that our proposed F-DRL can converge faster and obtain higher rewards with smaller fluctuations in the training process. 
	
	Fig. \ref{fig5} demonstrates the trajectory of robots versus different $M_{\rm R}$, where the performance of QoS-based energy efficiency (EE) policy and QoE-based EE policy is compared.
	The parameters are set as $N_{\rm R} = 4, N_{\rm P} = 6, N = 1$ and $K = 3$.
	The background in Fig. \ref{fig5} reflects the communication quality of downlink channels. 
	As expected, we find that the RIS enhances channel conditions, especially alleviating the severe signal strength degradation caused by the walls.
	The QoS-based EE policy maintains better channel conditions rather than Baseline \ref{baseline3}, especially when $M_{\rm R} > 20$.
	It is because that Baseline \ref{baseline3} cares more about the bad channel coefficients, while QoS-based EE policy cares more about the sum of channel conditions.
	Moreover, the result shows that QoS-based EE policy in the considered system can achieve higher energy efficiency, while the robot with the worst channel condition always maintains a required data rate in NOMA-based systems under Baseline \ref{baseline3}, because the logarithmic function is more sensitive to small data rate changes.
	
	In Fig. \ref{fig4}, the energy efficiency under different environmental parameters is illustrated.
	When $N_{\rm R} = 4$ and $N_{\rm P} = 6$, the energy efficiency is evaluated versus $P_{\max}$ by changing the number of robots $K$, multiple access technologies  $z \in \{ \rm{NOMA,OMA}\} $, and the number of RIS elements $M_{\rm R}$.
	We find that RIS is helpful to obtain higher energy efficiency.
	This is mainly because that the RIS can overcome signal blockage by adjusting the radio environment.
	Meanwhile, when $K = 3$, energy efficiency significantly increases with $M_{\rm R}$, and maintains smaller improvement with $ 20 \le M_{\rm R} \le 30$.
	Nevertheless, the energy efficiency increases with $0 \le M_{\rm R} \le 30$ when $K = 4$.
	Such phenomenon reveals that there exists the suitable transmit power budget $P_{\max}$ and RIS elements $M_{\rm R}$ satisfying the communication demands with lower values.
	Moreover, NOMA-RIS-based system gains higher energy efficiency than OMA-RIS-based benchmarks, because NOMA signals are superimposed in the same time-frequency resources and obtains enhanced bandwidth efficiency.
	In addition, fewer robots and smaller $P_{\max}$ lead to lower energy efficiency.

	\section{Conclusion}
	
	We studied a long-term energy efficiency maximization problem of RIS-assisted indoor multi-robot systems.
	Through training agents in a semi-distributed manner, we developed a novel methodology for the communication-aware design problem by controlling the trajectory and downlink transmit power at local robots, and designing the RIS phase shifts at the AP.
	Owing to the decentralized nature of the proposed F-DRL, the dynamics in such a multi-robot system can be well handled.
	Numerical simulations demonstrated that our designed F-DRL converges faster than the centralized method
	%and the NOMA-RIS-based schemes {can} achieve higher energy efficiency than OMA-RIS-based {schemes}.
	and adapts to the changes in the number of robots, while maintaining high performance in NOMA-RIS design.

	\bibliographystyle{IEEEtran}
	\bibliography{IEEEabrv,ref}
	
\end{document}